\documentclass[11pt,a4paper]{article}
\usepackage{times,latexsym}
\usepackage{url}
\usepackage[T1]{fontenc}
\usepackage{array,ragged2e}
\usepackage{textgreek}
\usepackage{makecell}
\usepackage{multirow}
\usepackage{booktabs}
\usepackage{caption}
\usepackage{subcaption}
\usepackage{comment}
\usepackage{graphicx}
\usepackage[acceptedWithA]{tacl2021v1}

\usepackage{tacl2021v1}

\usepackage{xspace,mfirstuc,tabulary}

\newif\iftaclinstructions
\taclinstructionsfalse 
\iftaclinstructions
\renewcommand{\confidential}{}
\renewcommand{\anonsubtext}{(No author info supplied here, for consistency with
TACL-submission anonymization requirements)}
\newcommand{\instr}
\fi

\iftaclpubformat 

\else

\fi

\title{Can large language models generalize analogy solving like children can?}

\author{
   Claire E. Stevenson$^\diamond$,
   Alexandra Pafford$^\diamond$,
   Han L. J. van der Maas$^\diamond$
   \&
   Melanie Mitchell$^\dagger$ \\
   $^\diamond$Psychological Methods, University of Amsterdam, the Netherlands \\
   $^\dagger$Sante Fe Institute, USA \ \\
   \texttt{c.e.stevenson@uva.nl, h.l.j.vandermaas@uva.nl, mm@santafe.edu}
 }

\author{Claire E. Stevenson\textsuperscript{1} \and Alexandra Pafford \textsuperscript{1} \and Han L. J. van der Maas \textsuperscript{1,2} \and Melanie Mitchell\textsuperscript{2} \\
  \textsuperscript{1} Psychological Methods, University of Amsterdam, the Netherlands\\
  \textsuperscript{2} Sante Fe Institute for Complexity, Sante Fe, AZ, USA \\
  \texttt{c.e.stevenson@uva.nl, h.l.j.vandermaas@uva.nl, mm@santafe.edu}\\}
\date{}

\begin{document}

\maketitle
\begin{abstract}
In people, the ability to solve analogies such as ``body : feet :: table : ?'' emerges in childhood, and appears to transfer easily to other domains, such as the visual domain ``$($~:~$)$~::~$<$~:~?''. Recent research shows that large language models (LLMs) can solve various forms of analogies. However, can LLMs generalize analogy solving to other domains like people can? To investigate this, we had children, adults, and LLMs solve a series of letter-string analogies (e.g.,~a~b~:~a~c~::~j~k~:~?) in the Latin alphabet, in a near transfer domain (Greek alphabet), and a far transfer domain (list of symbols). Children and adults easily generalized their knowledge to unfamiliar domains, whereas LLMs did not. This key difference between human and AI performance is evidence that these LLMs still struggle with robust human-like analogical transfer.
\end{abstract}

\section{Introduction}
You may be familiar with the analogy ``consciousness is like an iceberg''. Here, people intuitively infer the below-the-surface depth and complexity of consciousness by relating it to an iceberg, whose mass is mostly found under water, just as our subconscious dwells under our conscious minds. This ability emerges in childhood \citep{goddu2020, gentner1988, stevenson2018}. However, it is a subject of debate whether analogical reasoning has emerged in Large Language Models (LLMs) \citep{webb2023emergent, lewis2025evaluating, hodel2023response, webb2024evidence}. More importantly, are LLMs able to solve analogies at this level of conceptual abstraction and generalize to novel domains \citep{mitchell2021abstraction, shiffrin2023probing}? In this study, we investigate analogical transfer at two levels of abstraction (near and far), and compare LLM performance not only to adults, but also to children, who are still developing analogical reasoning abilities. We ask the question: Can LLMs can generalize analogy solving like children can?

\begin{figure}[t]
    \centering
    \includegraphics[width=\linewidth]{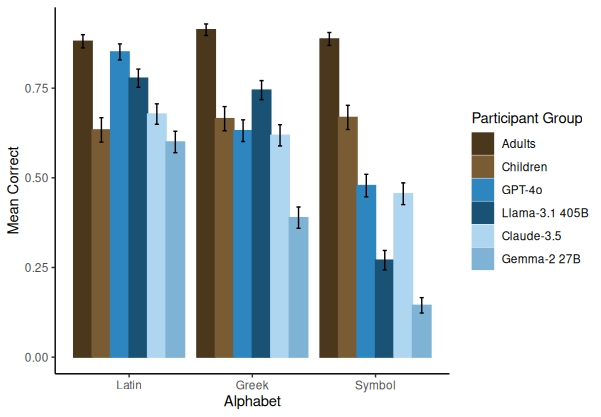}
    \caption{{\fontsize{10}{12}\selectfont Human vs LLM performance on letter-string analogies across alphabet domains.}}
    \label{fig:letstr_main_result}
\end{figure}

Analogical reasoning, the process of applying a known concept to understand something new through relational similarity, is fundamental to the way people think and learn \citep{holyoak2012analogy, gentner2017analogy}. This is because we humans can easily generalize \textemdash that is, transfer principles discovered in one domain to new domains that share varying degrees of similarity with the original \citep{doumas2022theory}. This can be principles in near contexts that are similar in terms of concrete attributes (e.g., shape, ``a pyramid is like an iceberg'') or in farther contexts that are only similar in terms of abstract relations (e.g., abstraction of depth, ``consciousness is like an iceberg'') \citep{barnett2002and}. Near analogies tend to be easier for both adults and children to solve than far analogies \citep{johnson2025verbal, jones2022, thibaut2016analogical}. And, in general, adults are better at solving analogies than children. But, when the required domain knowledge and a causal framing are present, then children can solve analogies such as ``body is to feet as table is to ?'' as early as the 3-4 years-old \citep[e.g.,][]{goddu2020, goswami1991analogical}. And when analogies are presented in a more challenging or far context, young children tend to revert to associative strategies, e.g., replying `egg' to `dog is to doghouse as chicken is to ?' instead of `chicken coop' \citep{stevenson2018, gentner1988, thibaut2016analogical}.

There are many tasks used to study analogical reasoning and transfer in people, from verbal to geometric to scene analogy problems \citep[e.g.,][]{ichien2020, richland2006children, mulholland1980components}. However, many of these tasks are either not suitable for children (e.g., verbal analogies may contain unfamiliar words or relations for children) or to LLMs (e.g., visual analogies designed for children are still difficult for today's multimodal models \citet{yiu2024kiva}). Therefore, we need a domain that is text-based, but doesn't require domain knowledge beyond what a typical child or LLM would know.  Letter-string analogies fit the bill as they require very little domain knowledge and offer an idealized scenario to examine analogical reasoning in a “pure, uncontaminated way” \citep[][ p. 3]{hofstadter1984}. In these puzzles, a string of letters is transformed according to one or more rules, and the task is to use analogy and apply the same transformations to a new string. For example, ``If abc changes to abd, what should pqr change to?'' \citep{mitchell2021abstraction}. 

Letter-string analogy solving has been studied in human adults and LLMs. For example, \citet{webb2023emergent} showed that GPT-3 is able to solve letter-string analogies better than college students. \citet{lewis2025evaluating} showed that GPT-models solved letter-string analogies at about 60\% accuracy in the Latin alphabet domain, somewhat below the level of adults they tested. Interestingly, \citet{lewis2025evaluating} and \citet{hodel2023response} found that GPT-3's performance degraded when presented with these same analogies using an alphabet of shuffled letters. Moreover, \citet{lewis2025evaluating} showed that GPT-models had great difficulty solving letter-string analogies in an unfamiliar alphabet of symbols, whereas people did not. As such, there is conflicting evidence of whether LLMs can generalize analogy solving to novel domains \citep{lewis2025evaluating, webb2024evidence, hodel2023response}, something that comes easily to adults \citep[e.g.,][]{thibaut2022acrossformats, doumas2022theory}, and that even children appear capable of when domains share structural similarities \citep{chen1996children, gentner1986systematicity, bobrowicz2020flexibility, holyoak1984development}. Thus, while there is some evidence to suggest that LLMs can solve letter-string analogies at around the same level as people, it is unclear whether these models understand the problem and are actually using analogical reasoning \citep{opielka2024large, johnson2025verbal, moskvichev23}. 

In this study, we investigate whether LLMs can generalize analogy solving to new domains like adults and 8-year-old children can at two levels of abstraction. To this end, we compare how adults, children, and LLMs generalize analogy solving on the letter-string task to both near (Greek alphabet) and far (Symbol list) domains. 

\section{Method}
We compared 42 children (7-9 year-olds), 62 adults, and 55 runs of each of four LLMs (Claude-3.5 \citet{llm-claude3}; Gemma-2 27B \citet{llm_gemma}; GPT-4o \citet{llm_gpt4}; and Llama-3.1 405B \citet{llm_llama2}) on a set of letter-string analogies under three alphabet conditions: Latin, Greek and a Symbol list. 

\subsection{Materials}
\subsubsection{Letter-String Analogy Task} 
Letter-string analogies, pioneered by Hofstadter \citeyear{hofstadter1984}, are a type of proportional analogy (A is to B as C is to D) involving alphabetic strings. For example, ``If the string of letters \textit{abc} changes to \textit{abd}. How would you change the string \textit{pqrs} in the `same way'?'' \citep{mitchell2021abstraction}. Such letter-string analogies can be solved in multiple ways. For example, shifting the last letter and responding \textit{pqrt} is what people tend to prefer (and what we consider ``correct'' in this context).  However, another possible solution could be \textit{pqrd}, where a literal rule is applied, namely replacing the last letter with \textit{d}.


\paragraph{Rules} There are several possible transformations from A to B and generalizations from A to C as described in \citet{webb2023emergent}. We use only the simplest transformations of successor (one and two after), predecessor (one before) and repetition, and the generalizations are limited to shifting in the alphabet and letter repetitions \textemdash rules that children are expected to be familiar with. 

\begin{figure}
    \centering
    \begin{subfigure}{.3\textwidth}
        \centering
        \includegraphics[width=.95\linewidth]{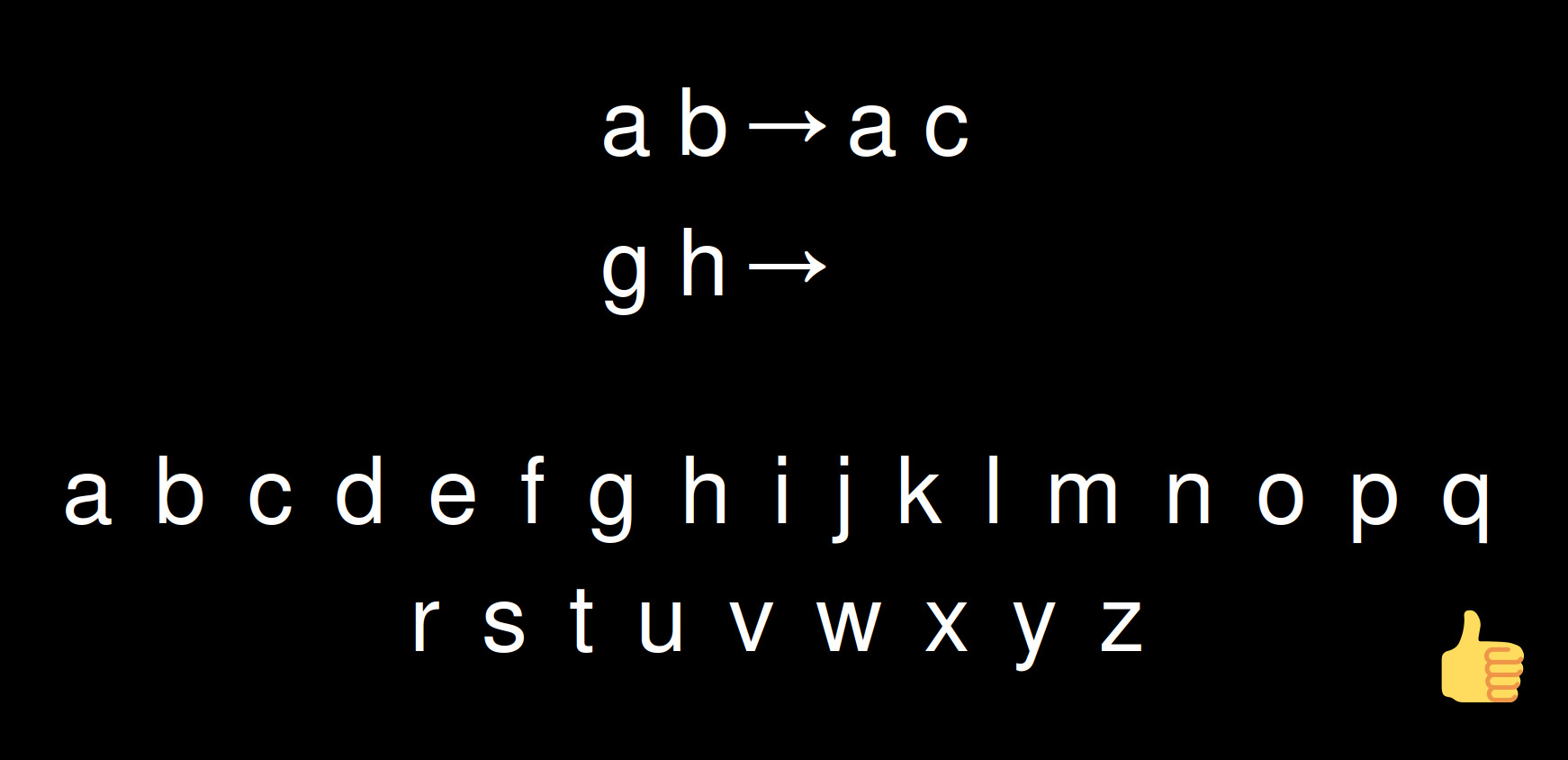}
        \caption{Latin}
    \end{subfigure}
    \begin{subfigure}{.3\textwidth}
        \centering
        \includegraphics[width=.95\linewidth]{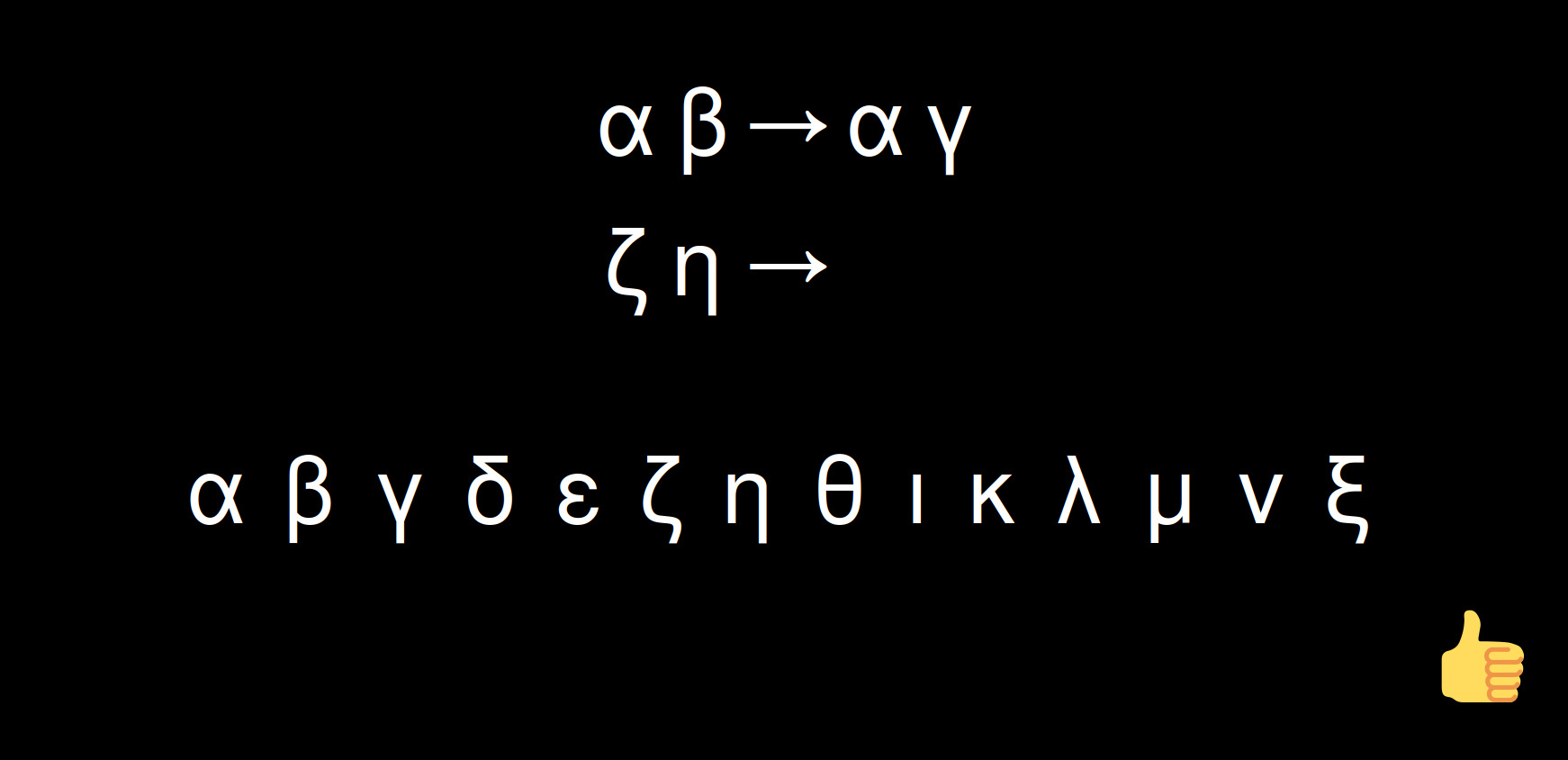}
        \caption{Greek}
    \end{subfigure}
    \begin{subfigure}{.3\textwidth}
        \centering
        \includegraphics[width=.95\linewidth]{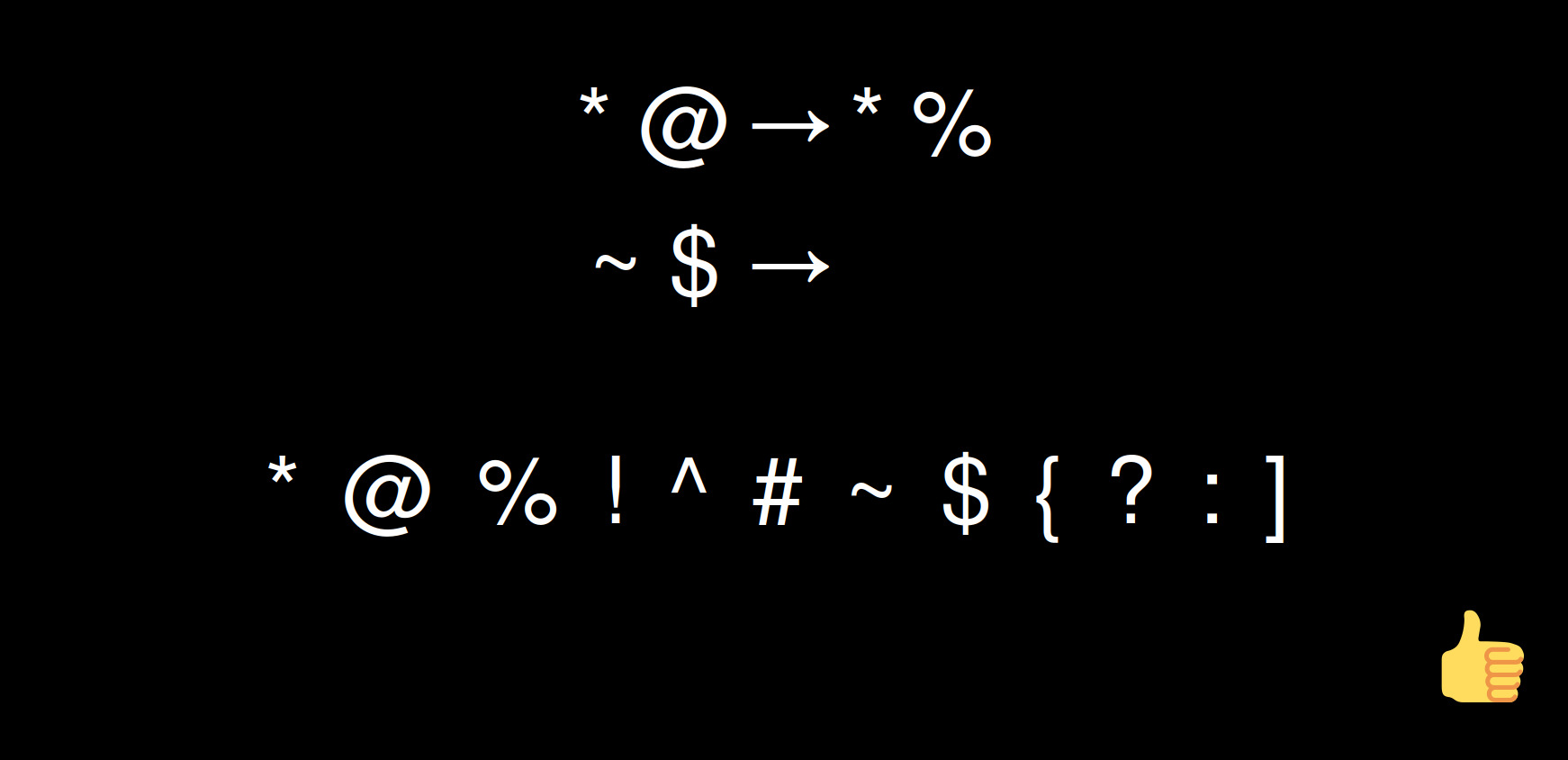}
        \caption{Symbol}
    \end{subfigure}
    \caption{Letter-string analogy task item 1 in (a) baseline alphabet: Latin, (b) near transfer alphabet: Greek, and (c) far transfer alphabet: Symbol.}
    \label{fig:item1}
\end{figure}

\paragraph{Alphabets} \label{alphabets} For each of the items in the Latin alphabet we also created a near transfer version using the Greek alphabet and a far transfer version in our invented Symbol alphabet: \textbf{\textasteriskcentered{} @ \% ! \textasciicircum{} \# \textasciitilde{} \$ \{ ? = :} (see Figure \ref{fig:item1} for an example). We chose the Greek alphabet as near transfer domain because Greek symbols are somewhat visually similar to the Latin alphabet, but otherwise unfamiliar to the children in our study. We presented actual Greek symbols to humans, but chose the written version (i.e., alpha, beta, etc.) for LLMs based on their ability to list the Greek alphabet in this form upon request. We chose to use an ordered list of Symbols for far transfer, because it is an unfamiliar `alphabet' that neither people nor the LLMs had seen before in this context, but at the same time were both able to process (i.e., the children can identify differences visually and for the LLMs these are common symbol keys on a keyboard). The constructed items for each alphabet were kept consistent, where the same transformations and generalizations from item 1 of the Latin alphabet were also used for item 1 of the Greek and Symbol alphabets. See Table \ref{tbl:itemset} in 
the Results section for an overview of all items.

\subsubsection{Human Data Collection}
\paragraph{Procedure} Both children and adults completed the task in a browser. They were first shown the Latin alphabet and told that they would solve puzzles with these. For adults there was a simple example with feedback as the study was carried out fully online. For children, the interface was explained and demonstrated in person. Participants then solved two simple practice items without feedback (used to ensure understanding of task). Then for each alphabet, they were shown the list of letters/symbols and told they would again solve puzzles using these letters/symbols, where the Greek and Symbol alphabets were referred to as ``secret code'' letters for children. There were five items for each alphabet, with 15 items total. 

\paragraph{Adults} We collected adult data online from fluent English speakers through the Prolific research participant recruitment platform. We recruited 68 adults of 18 years or older (M=24.0, SD =7.33, 50\% female) who had completed secondary education or higher and resided in the Netherlands or neighboring countries (as children were recruited in the Netherlands). We also required that they have no language disorders and have (corrected-to-) normal vision to ensure they could see/process the task, that they use a device at least 2x a week (to ensure digital fluency), and that they have a 95\% or higher approval rating on Prolific to ensure high quality data from the participants. Based on the pre-registered exclusion criteria for adults (answering >80\% of items), 6 adults were excluded. 

\paragraph{Children} Data was collected from 44 children (7-9 year-olds, M=8.26, SD=0.67) at a local school on an electronic tablet. The recruited school is a public Montessori school and emphasizes natural materials and does not use tablets or computers in this age group. All children from the participating classrooms were included in the study, as language disorders are generally not yet assessed in this age group in the Netherlands. The researchers gave spoken instructions given the limited reading abilities in this age group. The children then completed the task independently. We excluded two children, because they did not complete the task. 

\subsubsection{LLM Data Collection} 
We collected data from LLMs from four types of models: Anthropic's Claude-3 and Claude-3.5; Google's Gemma-2-9B and Gemma-2 27B; Open AI's GPT-3, GPT-3.5, GPT-4, and GPT-4o; Meta's Llama-3.1-8B, Llama-3.1-70B, Llama-3.1-405B. For each model type, the newest and largest model had the best performance. Therefore, to provide clear and concise results our main results comparing human and LLM performance report on this selected set of models: Anthropic's Claude-3.5, Google's Gemma-2 27B, Open AI's GPT-4o, and Meta's Llama-3.1 405B. A brief overview of the results of other models can be found in Section \ref{sec:scale}; the full dataset is available in our GitHub repository.

\paragraph{Procedure}
We presented the analogies in chat completion mode using Python APIs from Anthropic for Claude models, from Open AI for GPT models and from Together AI for the remaining models, which are all open source. We specified a temperature of 0 for near-deterministic data collection and set the maximum number of tokens to 10. 

\paragraph{Prompt}
\label{meth:prompt}
The LLM general instruction was as follows: \texttt{We are going to do puzzles with the letters or symbols `[Latin alphabet|Greek alphabet|Symbol list]'. Example 'if a changes to b, then j changes to k'.}

Per item the LLMs received the instruction and item as follows: \texttt{The [letter|symbol] list is `[Latin alphabet|Greek alphabet|Symbol list]'. If [A] changes to [B], what does [C] change to ?}.

\paragraph{Pre-pending previous conversation}
\label{meth:prepend}
Also, following \cite{webb2023emergent}'s approach to administering verbal analogies and digit matrices, all previous conversation with the LLM was pre-pended to each successive item so that the models could learn while testing just as people could. This seemed especially important because the exact same items with the same rules were applied in the same order from one alphabet set to the next. We also ran the tasks without pre-pending previous conversation, which generally resulted in lower LLM performance (see Appendix \ref{app:noprevmsg}).

\paragraph{Prompting templates} We tested 5 different prompt templates for presenting items to LLMs, as prompt engineering can change the LLMs' performance on the task. Results are reported for the best performing template as shown above: \texttt{If {A} changes to {B}, what does {C} change to?}. See the Appendix \ref{tbl:templates} for more details on the different templates and results.

\paragraph{Differences between Human and LLM Procedures}
To keep the conditions for the LLM data collection similar to that of people and, especially to fairly compare LLM results to those of children, we presented all analogies in a zero-shot setting using the same instructions that we spoke to the children. There were two exceptions. First, with children we referred to the Greek and Symbol alphabets as ``secret code letters'', whereas this was `alphabet' and `(ordered) list' respectively for adults and LLMs. Second, the LLMs received the worked example 'if a changes to b, then j changes to k' that humans did not receive.

\paragraph{Item variants for LLMs}
\label{itemvariants} 
To enable robust comparisons between individual LLMs and groups of people, we adopted a similar methodology to Webb et al. \citeyear{webb2023emergent} and administered approximately as many variants of the task to each LLM as we had people who solved it. To do so, we created variants of each item by systematically shifting all of the characters in the item. For example, ``a b : a c :: l m : ?'' became ``b c : b d :: m n : ?''. For each of the 5 items per alphabet administered to humans (see Table \ref{tbl:itemset}), we created 4 item variants, i.e., shifted 1-4 elements to the left and/or right. We then systematically selected item variants to create 55 unique parallel testlets (required number based on power analysis from pre-registration; Note: we administered each testlet as a `conversation' containing multiple messages to recreate how we tested humans, see \ref{meth:prepend}). This allowed us to have robust estimates of LLM performance, while creating some variation in LLM data and enabling us to compute SE's for statistical analyses. 

\begin{table*}[ht]
\centering
\caption{{\fontsize{10}{12}\selectfont Base item set administered to adults, children, and LLMs.}}
\label{tbl:itemset}
{\fontsize{10}{12}\selectfont
\begin{tabular}{cccccccc}
\toprule
\textbf{Item ID} & \textbf{Alphabet} & \textbf{A} & \textbf{B} & \textbf{C} & \textbf{D} & \textbf{AB Rule} & \textbf{AC Rule} \\
\midrule
A & Practice & a & b & j & k & successor\_1 & shift \\
B & Practice & c d & c d d & j k & j k k & repeat\_1 & shift \\
\midrule
1 & Latin & a b & a c & g h & g i & successor\_1 & shift \\
2 & Latin & c d & c c e e & m n & m m o o & successor\_1, repeat\_2 & shift \\
3 & Latin & e f & e h & k l & k n & successor\_2 & shift \\
4 & Latin & d e & d f f & g h & g i i & successor\_1, repeat\_1 & shift \\
5 & Latin & c d & b d & m m n n & l l n n & predecessor\_1 & shift, repeat\_2 \\
\midrule
1 & Greek & \textalpha\ \textbeta & \textalpha\ \textgamma & \textzeta\ \texteta & \textzeta\ \texttheta & successor\_1 & shift \\
2 & Greek & \textgamma\ \textdelta & \textgamma\ \textgamma\ \textepsilon\ \textepsilon & \textkappa\ \textlambda & \textkappa\ \textkappa\ \textmu\ \textmu & successor\_1, repeat\_2 & shift \\
3 & Greek & \textepsilon\ \textzeta & \textepsilon\ \texttheta & \textiota\ \textkappa & \textiota\ \textmu & successor\_2 & shift \\
4 & Greek & \texteta\ \texttheta & \texteta\ \textiota\ \textiota & \textlambda\ \textmu & \textlambda\ \textnu\ \textnu & successor\_1, repeat\_1 & shift \\
5 & Greek & \textbeta\ \textgamma & \textalpha\ \textgamma & \textnu\ \textnu\ \textxi\ \textxi & \textmu\ \textmu\ \textxi\ \textxi & predecessor\_1 & shift, repeat\_2 \\
\midrule
1 & Symbol & * @ & * \% & \textasciitilde\ \$ & \textasciitilde\ \{ & successor\_1 & shift \\
2 & Symbol & \% ! & \% \% \^{} \^{} & = : & = = ) ) & successor\_1, repeat\_2 & shift \\
3 & Symbol & @ \% & @ \^{} & \# \textasciitilde & \# \{ & successor\_1 & shift \\
4 & Symbol & ! \^{} & ! \# \# & \$ \{ & \$ = = & successor\_1, repeat\_1 & shift \\
5 & Symbol & \^{} \# & ! \# & = = : : & \{ \{ : : & predecessor\_1 & shift, repeat\_2 \\
\bottomrule
\end{tabular}}
\end{table*}

\section{Results}
We use mixed ANOVAs to (1) compare performance between our between-subjects participant groups (Adults, Children, and each of the LLMs) on the Latin alphabet and (2) test whether each participant group could generalize analogy solving by performing similarly across alphabets (i.e., our repeated within-subjects factor). All plots show the  means and standard errors as error bars.

\subsection{RQ1: How well do LLMs solve letter-string analogy problems in the Latin alphabet compared to adults and children?}
We expected LLMs to be able to solve letter-string analogies with the Latin alphabet at the same level as adults \citep{webb2023emergent} and that both adults and LLMs would outperform children \citep{thibaut2016analogical} (hypotheses H1a-c). Similar to what we expected, adults and some LLMs, except Google's Gemma-2 27B and Anthropic's Claude 3.5, performed better than children in the Latin alphabet domain. Open AI's GPT-4o performed similarly to adults, followed closely by Meta's Llama-3.1 405B. See Figure \ref{fig:letstr_main_result} and Table \ref{tbl:mainresults} for more detailed results.

\begin{table*}
\centering
{\fontsize{10}{12}\selectfont
\begin{tabular}{lccccccccc}
\toprule
\textbf{Participant Group} & \textbf{n} & \multicolumn{2}{c}{\textbf{Latin}} & \multicolumn{2}{c}{\textbf{Greek}} & \multicolumn{2}{c}{\textbf{Symbol}} \\
\cmidrule(lr){3-4} \cmidrule(lr){5-6} \cmidrule(lr){7-8}
 & & \textbf{Mean} & \textbf{SD} & \textbf{Mean} & \textbf{SD} & \textbf{Mean} & \textbf{SD} \\
\midrule
Adults         & 62 & 0.88 & 0.16 & 0.91 & 0.13 & 0.89 & 0.23 \\
Children       & 41 & 0.62 & 0.22 & 0.66 & 0.23 & 0.67 & 0.30 \\
Claude-3.5     & 54 & 0.68 & 0.18 & 0.62 & 0.21 & 0.46 & 0.24 \\
Gemma-2 27B    & 54 & 0.60 & 0.24 & 0.39 & 0.20 & 0.14 & 0.15 \\
GPT-4o         & 54 & 0.85 & 0.18 & 0.63 & 0.21 & 0.48 & 0.18 \\
Llama-3.1 405B & 54 & 0.79 & 0.16 & 0.74 & 0.19 & 0.27 & 0.20 \\
\bottomrule
\end{tabular}}
\caption{\fontsize{10}{12}\selectfont Descriptive statistics on letter-string analogy performance by Participant Group and Alphabet.}
\label{tbl:mainresults}
\end{table*}

\subsection{RQ2: How well do adults, children and LLMs generalize letter-string analogy solving from Latin to Greek (near) and Symbol (far) alphabets?}

As expected, adults and children performed similarly across alphabets (see Figure \ref{fig:letstr_main_result}). But, as we suspected, LLM performance indeed degraded in less familiar alphabets (ANOVA results shown in Table \ref{tbl:posthocalphabeteffect}). More specifically, for each model, performance degraded significantly from the Latin to Greek alphabet (posthoc Bonferonni-corrected t-test results all \textit{p}<.001, except for Llama-3.1 405B\textit{p} = 0.012) and then again from the Greek alphabet to the Symbol list (posthoc Bonferonni-corrected t-test results all \textit{p}<.001).

\begin{table*}
\centering
\caption{{\fontsize{10}{12}\selectfont Post hoc ANOVA Results for main Alphabet effect by Participant Group}}
\label{tbl:posthocalphabeteffect}
{\fontsize{10}{12}\selectfont
\begin{tabular}{lccccc}
\toprule
\textbf{Participant Group} & \textbf{Effect} & \textbf{DFn} & \textbf{DFd} & \textbf{F} & 
\textbf{p Adjusted} \\
\midrule
Adults         & 1.59 & 96.9 & 0.95 & 1.000 \\
Children       & 2.00 & 76.0 & 0.27 & 1.000 \\
Claude-3.5     & 1.65 & 87.6 & 29.5 & <.001 \\
Gemma-2 27B    & 2.00 & 106.0 & 88.2 & <.001 \\
GPT-4o         & 2.00 & 100.0 & 55.0 & <.001 \\
Llama-3.1 405B & 1.70 & 90.1 & 135.0 & <.001 \\
\bottomrule
\end{tabular}}
\end{table*}

\subsection{RQ3: Why can't LLMs generalize letter-string analogy solving like children?}
\subsubsection{Performance by Item}
To understand why the LLM's had trouble generalizing letter-string analogy solving we examined item-by-item performance.
Table \ref{tbl:performance_by_item} shows an overview. Here we see that the LLMs and humans perform best on item 1, that involves only the first successor transformation, and worst on item 5, that involves both the predecessor transformation and repetition generalization. Item 2 also involves the same repetition rule as item 5, but was solved better by LLMs and children; therefore, it appears that the predecessor rule is what gives both LLMs and children the most trouble. The other item people and LLMs have relatively more trouble with is item 3. This item involves the second successor rule. In sum, the predecessor and second successor rules appear to be the most difficult rules from our item set for people and LLMs to apply.  
\begin{table*}
\centering
\caption{{\fontsize{10}{12}\selectfont Mean proportion correct (SD) by Participant Group for each Item}}
\label{tbl:performance_by_item}
{\fontsize{10}{12}\selectfont
\begin{tabular}{lccccc}
\toprule
 & & & \textbf{Item} &  & \\
\textbf{Participant Group} & \textbf{1} & \textbf{2} & \textbf{3} & \textbf{4} & \textbf{5} \\
\midrule
Adults         & 0.97 (0.18) & 0.94 (0.25) & 0.82 (0.39) & 0.94 (0.24) & 0.81 (0.40) \\
Children       & 0.85 (0.36) & 0.75 (0.44) & 0.52 (0.50) & 0.76 (0.43) & 0.38 (0.49) \\
Claude-3.5     & 0.90 (0.30) & 0.65 (0.48) & 0.54 (0.50) & 0.62 (0.49) & 0.20 (0.40) \\
Gemma-2 27B    & 0.62 (0.49) & 0.27 (0.45) & 0.38 (0.49) & 0.37 (0.48) & 0.25 (0.43) \\
GPT-4o         & 0.92 (0.27) & 0.73 (0.45) & 0.45 (0.50) & 0.78 (0.41) & 0.39 (0.49) \\
Llama-3.1 405B & 0.83 (0.37) & 0.62 (0.49) & 0.53 (0.50) & 0.57 (0.50) & 0.44 (0.50) \\
\bottomrule
\end{tabular}}
\end{table*}

\subsubsection{Next-Previous Letter Task}
We designed the Next-Previous Letter Task to check that the LLMs had the requisite knowledge of predecessor and successor to solve letter-string analogies. For this new task we provided an ordered list of letters/symbols and asked the LLMs what the previous and next two letters were given a specific letter. Each rule was tested 5 times, resulting in 20 items total.

We used this optimized prompt: \texttt{Here is an ordered list of letters or symbols [Latin alphabet|Greek alphabet|Symbol list]. Which letter or symbol is [one|two] [before|after] [X]?}. See Table \ref{tbl:nextprev_items} for rules and example items.

\begin{table}[ht]
    \centering
    \caption{{\fontsize{10}{12}\selectfont Next-Previous Letter Task: Example Items From the Latin Alphabet.}}
    \label{tbl:nextprev_items}
    {\fontsize{10}{12}\selectfont
    \begin{tabular}{ccc}
        \toprule
        \textbf{X} & \textbf{correct} & \textbf{Rule} \\
        \midrule
        c & d & next\_1 \\
        c & e & next\_2 \\
        d & c & prev\_1 \\
        e & c & prev\_2 \\
        \bottomrule
    \end{tabular}}
\end{table}

\begin{figure}
    \centering
    \includegraphics[width=\linewidth]{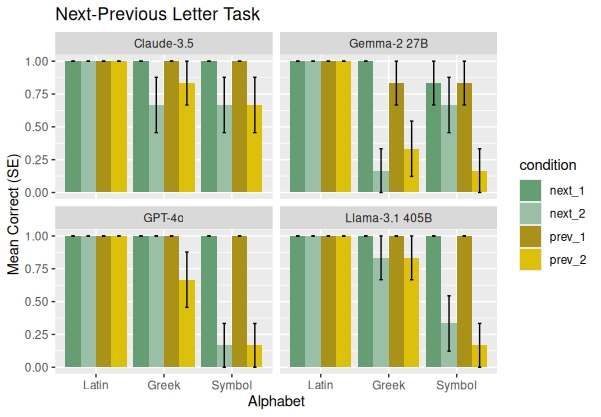}
    \caption{{\fontsize{10}{12}\selectfont LLM performance by rule type across alphabet domains.}}
    \label{fig:prevnext_result}
\end{figure}

As can be seen in Figure \ref{fig:prevnext_result}, all models do best when asked to identify the next or previous letter and worse when it concerns identifying two before or two after. Furthermore, Claude-3.5 performed well and similarly in all three domains, which is in contrast to its letter-string analogy performance that degrades from baseline to near to far domains. Similarly, GPT-4o performs well on the Next-Previous task in the Latin and Greek domain, but in the analogy task, its performance degrades from Latin to Greek. For Llama-3.1 405B transfer from the Latin to Greek to Symbol domain is similar across tasks, where in both tasks it does well with the Latin and Greek alphabets, but not the Symbol alphabet. Gemma-2 27B's performance is surprisingly more spotty here in the Greek domain than the Symbol domain. In conclusion, these results could explain why the LLMs have trouble with item 3, involving the second successor, but the results do not explain why they have trouble with item 5 involving the predecessor.

\subsubsection{Rule Check Task}
To better pinpoint why the LLMs had difficulty generalizing to other alphabets, we created a simplified version of the original analogy task that explicitly tested each rule in isolation. LLM prompts were exactly the same as in our original letter-string task, see Section \ref{meth:prompt}.

The rules were: (1) successor\_1, the next letter; (2) successor\_2, letter two places after; (3) predecessor\_1, the previous letter; (4) predecessor\_2, letter two places before; (5) repetition\_1, repeating the last letter and (6) repetition\_2, repeating both letters. Each rule was tested five times. See Table \ref{tbl:rulecheck_items} for examples.

\begin{table}[ht]
    \centering
    \caption{{\fontsize{10}{12}\selectfont Rule Check Task: Example Items From the Latin Alphabet.}}
    \label{tbl:rulecheck_items}
    {\fontsize{10}{12}\selectfont
    \begin{tabular}{ccccc}
        \toprule
        \textbf{A} & \textbf{B} & \textbf{C} & \textbf{D} & \textbf{Rule AB} \\
        \midrule
        c & d & h & i & successor\_1 \\
        c & e & h & j & successor\_2 \\
        d & c & h & g & predecessor\_1 \\
        e & c & h & f & predecessor\_2 \\
        c d & c d d & h i & h i i & repetition\_1 \\
        c d & c c d d & h i & h h i i & repetition\_2 \\
        \bottomrule
    \end{tabular}}
\end{table}

\begin{figure}
    \centering
    \includegraphics[width=\linewidth]{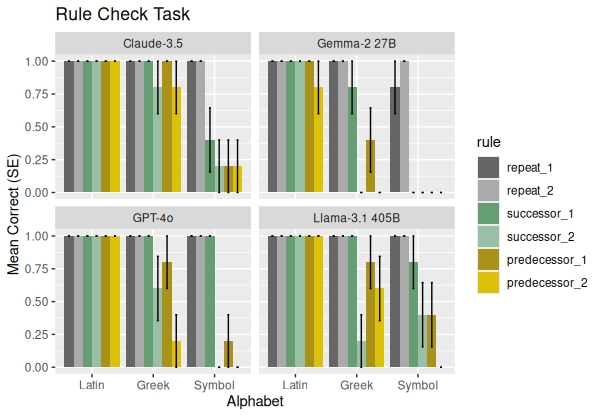}
    \caption{{\fontsize{10}{12}\selectfont Rule-specific performance across alphabet domains for LLMs.}}
    \label{fig:rulecheck_result}
\end{figure}

As Figure \ref{fig:rulecheck_result} shows, the LLMs we tested can solve all rules in the Latin alphabet and have no problem with repetition rules in the Greek and Symbol domains. The successor and predecessor rules were solved to differing degrees in the Greek alphabet, with Claude-3.5 performing best followed by GPT-4o. All models had trouble with the successor\_2 and predecessor rules in the Symbol alphabet, where only the successor\_1 rule sometimes formed an exception. This makes sense given the predict-the-next-token goal that LLMs are trained on \citep{mccoy2024embers}. 

\subsubsection{Error Analysis}
\label{sec:erroranalysis}
In general, when solving letter-string analogies there are often multiple rules that could underlie the change from A to B \cite{hofstadtermitchell1994}. Because we use very short strings, there are generally only two clearly correct responses. We considered the rules that people would generally prefer when responding, to be ``correct'', such as if \textbf{ab} changes to \textbf{ac}, then \textbf{gh} changes to \textbf{gi}. However, the literal rule of replacing the last letter with \textbf{c}, with response \textbf{gc} could also be considered correct.

\paragraph{Error Categories} To examine errors in more detail we created a set of categories based on those from \citep{lewis2025evaluating} and extended these to account for common errors in children \cite{stevenson2018}. In the \textbf{Literal rule} category, the change from A to B is literally copied to C such as \textbf{a~b~:~a~c~c~::~g~h~:~g~c~c} rather than providing the more common response of \textbf{g~i~i}. In the \textbf{One rule} category, the response is partially correct, but only (part of) one of the rules in the problem was applied, such as in responses to the previous example, \textbf{g~h~h} (only repetition applied) or \textbf{g~i} (only successor applied). Partially correct responses are common in children when problem load supersedes processing capacity \cite{stevenson2018}. In the \textbf{Incorrect rule} category, one of the other rules from our item set (i.e., successor, predecessor, repetition) was applied; for example, if the successor rule was used instead of the predecessor rule. For the \textbf{Copy rule}, the A, B or C term was copied as copying the C-term is common in young children \citep{stevenson2018, opielka2024large}. Finally, all remaining erroneous responses were placed in the \textbf{Other rule} category. Given that our task was less complex than in \citep{lewis2025evaluating} (i.e., shorter strings, fewer rules), we were able to automatically code these categories.

\begin{table*}
\centering
\caption{{\fontsize{10}{12}\selectfont Proportions of error categories by participant group. \textit{Note}. 5\% of children's responses were empty.}}
\label{tbl:erroranalysis}
{\fontsize{10}{12}\selectfont
\begin{tabular}{c c c c c c c c c c}
\toprule
\textbf{Participant Group} & \textbf{Correct} & \textbf{Literal Rule} & \textbf{One Rule} & \textbf{Incorrect Rule} & \textbf{Copy Rule} & \textbf{Other Rule} \\
\midrule
Adults         & 0.89 & 0.00 & 0.02 & 0.00 & 0.00 & \textbf{0.09} \\
Children       & 0.66 & 0.00 & 0.06 & 0.01 & 0.00 & \textbf{0.23} \\
Claude-3.5     & 0.58 & 0.05 & \textbf{0.19} & 0.08 & 0.01 & 0.09 \\
Gemma-2 27B    & 0.38 & \textbf{0.21} & 0.12 & 0.05 & 0.02 & \textbf{0.22} \\
GPT-4o         & 0.65 & \textbf{0.13} & 0.07 & 0.02 & 0.00 & \textbf{0.12} \\
Llama-3.1 405B & 0.60 & 0.08 & 0.10 & 0.02 & 0.00 & \textbf{0.20} \\
\bottomrule
\end{tabular}}
\end{table*}

Table \ref{tbl:erroranalysis} shows that adults and children did not use the Literal rule, whereas all models used it sometimes. For Gemma-2 27B and GPT-4o the Literal rule was one of the most common error types. The One rule was used most often in errors by Claude-3.5. The Incorrect and Copy rules were not used very often by people or models. And the Other rules were used most often by all, except Claude-3.5. 

\begin{figure}
    \centering
    \includegraphics[width=\linewidth]{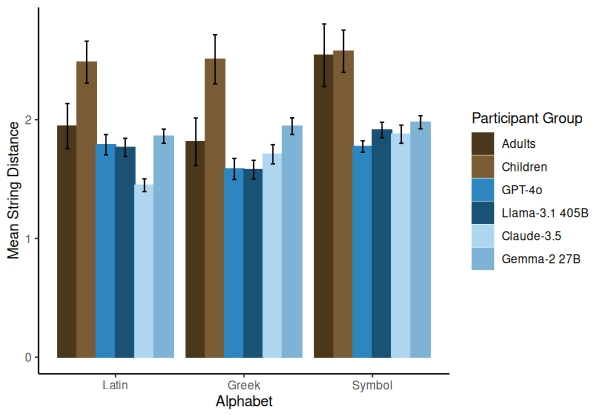}
    \caption{{\fontsize{10}{12}\selectfont Mean Levenshtein string distance between incorrect and expected responses.}}
    \label{fig:results_strdist}
\end{figure}

\paragraph{String Distance between ``Correct'' and ``Erroneous'' Responses} For each erroneous response we computed the Levenshtein string distance, also known as optimal string alignment distance, from the expected ``correct'' response to the given response. This distance counts the minimum number of edit operations (insertion, deletion, substitution) needed to change one string into the other. Here we investigate whether there are differences in mean Levenshtein distance between adults, children and LLMs for ``erroneous'' responses. Figure \ref{fig:results_strdist} shows that that the Levenshtein distance for ``erroneous'' responses is greater for children on all alphabets than for LLMs. For adults, this is only the case for the Symbol alphabet. For LLMs the Levenshtein distance hovers just under the 2 for all alphabets.  Note also that the standard errors for LLMs are also much smaller, but this is because the adults and sometimes children (Greek, Symbol alphabets) had far fewer ``erroneous'' responses to sample from. These results tells us that when children provide ``erroneous'' answers their responses tended to differ largely from the expected response. For example, three children responded `m m' to the item `If c d changes to b d, what does m m n n change to?', which has a Levenshtein distance of 6 from the expected response `l l n n'. The LLMs tended to provide 1 or 2 expected letters and 1 or 2 unexpected ones. For example, on the same item (and its variants) six GPT-4o runs provided `l m m n' as a response, with a Levenshtein distance of 2 from the expected response.

\section{RQ4: What is the effect of model size on letter-string analogy performance?}
\label{sec:scale}
\begin{figure*}
    \centering
    \begin{subfigure}{.24\textwidth}
        \centering
        \includegraphics[width=.95\linewidth]{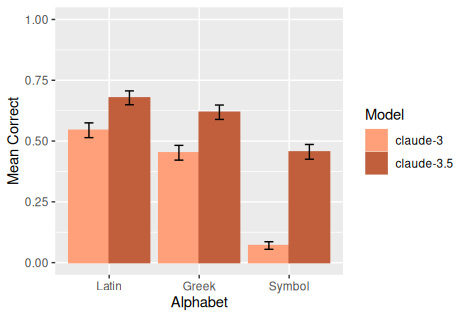}
        \caption{Claude}
    \end{subfigure}
    \begin{subfigure}{.24\textwidth}
        \centering
        \includegraphics[width=.95\linewidth]{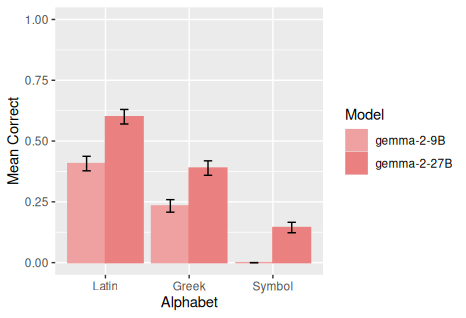}
        \caption{Gemma}
    \end{subfigure}
    \begin{subfigure}{.24\textwidth}
        \centering
        \includegraphics[width=.95\linewidth]{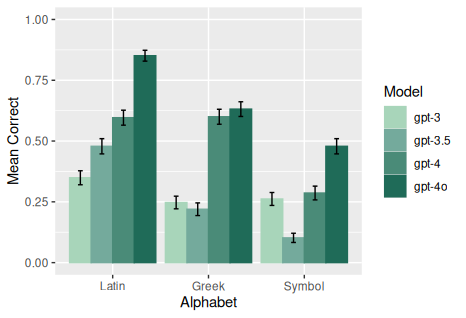}
        \caption{GPT}
    \end{subfigure}
        \begin{subfigure}{.24\textwidth}
        \centering
        \includegraphics[width=.95\linewidth]{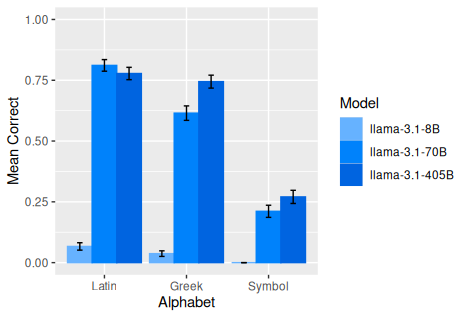}
        \caption{Llama}
    \end{subfigure}
    \caption{Effect of LLM size on proportion correct in letter-string analogy solving across alphabets Latin, Greek and Symbol.}
    \label{fig:scale}
\end{figure*}

In general, larger LLMs perform better on reasoning tasks than smaller LLMs \citep{wei2022emergent, huang2022reasoningllmssurvey}. As Figure \ref{fig:scale} shows, typical scaling laws generally appear to hold for how well LLMs generalize analogy solving in the letter-string domain. Especially in the Symbol domain, we observe a marked performance increase from smaller to larger models.

\section{Discussion}
Our main finding is that the LLMs we tested, using the same prompts given to children, were not able to generalize letter-string analogy solving like children can. LLMs perform at or above the level of children on letter-string analogies in the familiar Latin alphabet, but their performance on these same problems reduces somewhat when using the Greek alphabet (near transfer) and deteriorates greatly when using our Symbol alphabet (far transfer).

Why can't LLMs generalize when solving letter-string analogies? For some LLMs, this appears to be because they do not meet underlying requisites, such as indicating the predecessor or second successor of a letter in sequence. This would make sense given the predict-the-next-token goal that LLMs are trained on \cite{mccoy2024embers}. We tested this using the Next-Previous letter task, where models were explicitly prompted with an ordered list of letters or symbols and asked to identify the (second) successor or predecessor to a given letter or symbol. These results explain why the LLMs have trouble with analogies involving a second successor. But, the LLMs had little trouble identifying predecessors in the Next-Previous Letter task, so these results do not fully explain why LLM performance degrades from Latin to Greek and Symbol domains. 

The problem with LLMs transfer from the Latin to other domains seems to lie in that the conceptual abstraction of what constitutes an alphabet, such as being an ordered sequence, does not flexibly map to less familiar domains like it does in people. Evidence for this comes from the Rule Check task, where we tested LLMs on each rule in isolation. Here repetition rules could easily be applied to novel alphabets. This makes sense because repeating a character in a string can be done without knowing the alphabet. In contrast, LLMs had more trouble with predecessor and second successor rules. Both require an alphabet that is encoded as an ordered list of letters/symbols and an abstraction of what constitutes \textit{previous} and \textit{next}. This result aligns with previous work where GPT models could solve letter-string analogies with familiar alphabets in their standard order, but for shuffled alphabets performance dropped drastically, whereas for people performance remained the same \cite{hodel2023response, lewis2025evaluating}. We noted that in the Greek domain the letters were also ordered by unicode value, but in our Symbol domain they were not, which could perhaps explain why Greek items were easier. So, to check whether order was also a factor in our Symbol domain, we adapted the task to make the Symbol alphabet also ordered by their unicode values. However, this adaptation resulted only in some improvement in the Claude and Gemma models, and our findings still held (see Appendix \ref{app:orderedsymboltask}).

We also investigated which kinds of errors people and LLMs made. This is important because letter-string analogies, like many four-term visual analogies, apply ambiguous rules \cite[e.g.,][]{opielka2024large}, and can be solved correctly in multiple ways \cite{hofstadtermitchell1994}. The two main ways to solve the items in our task were what we considered the ``correct'' way (e.g., a~b~:~a~c~c~::~g~h~:~g~i~i) and the ``literal'' way (e.g., a~b~:~a~c~c~::~g~h~:~g~c~c). People did not use the ``literal'' rule, whereas the models all did to varying degrees (ranging from 5-21\%). The other main difference between human and LLM errors, was that children's erroneous responses were generally more distant (Levenshtein string distance) from the ``correct'' response than those of LLMs. This could be because children reverted to non-analogical strategies that we didn't account for in our error coding scheme, given that this is the first time letter-string analogies have been administered to children. 

Based on our results, LLMs appear unable to create on-the-fly representations of novel alphabets in the context of the letter-string analogies as well as the next-previous letter task \textemdash despite being given the ordered list of letters/symbols before each item. This inability was clear for both larger and smaller LLMs, although relative performance did scale with model size. A possible explanation lies in work studying the internal representations of LLMs, where abstract concepts  like "antonym" show invariant linear representations, but "previous" and "next" do not \cite{opielka2025analogical}. It appears that LLMs require an in-weight linear representation of an alphabet to successfully solve letter-string analogies. For novel alphabets, next-token-prediction does help them solve analogies with simple repetition and successor rules, but not with more complex rules and not at the level of children. Indeed, Webb et al. \citeyear{webb2024evidence} found that GPT-4 can only perform these abstractions by creating and executing code to map the novel alphabet to new positions and compute previous and next letters. This is of course very different from how children solve such problems. 

In contrast, our results show that in children, familiarity with letters or symbols does not influence letter-string analogy solving. As such, our results add to the accumulating evidence that questions whether reasoning actually occurs in these LLMs \citep{wu2024reasoning, gendron2024large, razeghi2022impact}. Interestingly, in \citeyear{schank1980aiintelligence}, Schank concluded that there wasn't much intelligence in artificial intelligence given its limited ability to generalize. Similarly, Doumas et al. (\citeyear{doumas2022theory}) argue that robust analogical transfer is a uniquely human ability. Based on our findings so far we concur, and now ask the question: Is generalization to unfamiliar domains indeed what separates human general intelligence from that of artificial general intelligence? The challenge now is to create uncontaminated far generalization tasks that AI models have not been trained on to answer this question.

\section{Acknowledgments}
This research was funded by the the Dutch Research Council (NWO) project "Learning to solve analogies: Why do children excel where AI models fail?" with project number 406.22.GO.029 awarded to Claire Stevenson. We thank Veerle Vijverberg and Talea Sibum for their assistance with data collection. We also thank the University of Amsterdam CreAI Lab for fruitful discussions related to this work and the anonymous reviewers and the Action Editor for their helpful comments.

\bibliographystyle{acl_natbib}
\bibliography{tacl_letstr}

\begin{thebibliography}{42}
\expandafter\ifx\csname natexlab\endcsname\relax\def\natexlab#1{#1}\fi

\bibitem[{Touvro~et al(2023)}]{llm_llama2}
H.~Touvro~et al. 2023.
\newblock \href {https://arxiv.org/pdf/2307.09288} {Llama 2: Open foundation and fine-tuned chat models}.
\newblock \emph{arXiv preprint arXiv:2307.09288}.

\bibitem[{Anthropic(2024)}]{llm-claude3}
Anthropic. 2024.
\newblock \href {https://www.anthropic.com/news/claude-3-family} {Introducing the next generation of claude}.
\newblock \emph{Anthropic preprint}.

\bibitem[{Barnett and Ceci(2002)}]{barnett2002and}
Susan~M Barnett and Stephen~J Ceci. 2002.
\newblock \href {http://rapunselshair.pbworks.com/f/barnett_2002.pdf} {When and where do we apply what we learn?: A taxonomy for far transfer.}
\newblock \emph{Psychological Bulletin}, 128(4):612.

\bibitem[{Bobrowicz et~al.(2020)Bobrowicz, Lindstr{\"o}m, Lindblom~Lov{\'e}n, and Psouni}]{bobrowicz2020flexibility}
Katarzyna Bobrowicz, Felicia Lindstr{\"o}m, Marcus Lindblom~Lov{\'e}n, and Elia Psouni. 2020.
\newblock \href {https://doi.org/10.3389/fpsyg.2020.573730} {Flexibility in problem solving: Analogical transfer of tool use in toddlers is immune to delay}.
\newblock \emph{Frontiers in Psychology}, 11:573730.

\bibitem[{Chen(1996)}]{chen1996children}
Zhe Chen. 1996.
\newblock Children's analogical problem solving: The effects of superficial, structural, and procedural similarity.
\newblock \emph{Journal of Experimental Child Psychology}, 62(3):410--431.

\bibitem[{Doumas et~al.(2022)Doumas, Puebla, Martin, and Hummel}]{doumas2022theory}
Leonidas~AA Doumas, Guillermo Puebla, Andrea~E Martin, and John~E Hummel. 2022.
\newblock \href {https://psycnet.apa.org/record/2022-26663-001} {A theory of relation learning and cross-domain generalization.}
\newblock \emph{Psychological Review}.

\bibitem[{GemmaTeam(2024)}]{llm_gemma}
GemmaTeam. 2024.
\newblock \href {https://arxiv.org/pdf/2403.08295} {Gemma: Open models based on gemini research and technology}.
\newblock \emph{arXiv preprint arXiv:2403.08295}.

\bibitem[{Gendron et~al.(2024)Gendron, Bao, Witbrock, and Dobbie}]{gendron2024large}
Gael Gendron, Qiming Bao, Michael Witbrock, and Gillian Dobbie. 2024.
\newblock Large language models are not strong abstract reasoners.
\newblock In \emph{Proceedings of the 33rd International Joint Conference on Artificial Intelligence}, IJCAI-24. International Joint Conferences on Artificial Intelligence Organization.

\bibitem[{Gentner(1988)}]{gentner1988}
D.~Gentner. 1988.
\newblock \href {https://doi.org/10.1111/j.1467-8624.1988.tb03194.x} {Metaphor as structure mapping: The relational shift}.
\newblock \emph{Child Development}, 59(1):47--59.

\bibitem[{Gentner and Hoyos(2017)}]{gentner2017analogy}
Dedre Gentner and Christian Hoyos. 2017.
\newblock \href {https://onlinelibrary.wiley.com/doi/full/10.1111/tops.12278} {Analogy and abstraction}.
\newblock \emph{Topics in Cognitive Science}, 9(3):672--693.

\bibitem[{Gentner and Toupin(1986)}]{gentner1986systematicity}
Dedre Gentner and Cecile Toupin. 1986.
\newblock \href {https://doi.org/10.1016/S0364-0213(86)80019-2} {Systematicity and surface similarity in the development of analogy}.
\newblock \emph{Cognitive science}, 10(3):277--300.

\bibitem[{Goddu et~al.(2020)Goddu, Lombrozo, and Gopnik}]{goddu2020}
M.~K. Goddu, T.~Lombrozo, and A.~Gopnik. 2020.
\newblock \href {https://doi.org/10.1111/cdev.13412} {Transformations and transfer: Preschool children understand abstract relations and reason analogically in a causal task}.
\newblock \emph{Child Development}, 91(6):1898--1915.

\bibitem[{Goswami(1991)}]{goswami1991analogical}
Usha Goswami. 1991.
\newblock Analogical reasoning: What develops? a review of research and theory.
\newblock \emph{Child development}, 62(1):1--22.

\bibitem[{Hodel and West(2023)}]{hodel2023response}
Damian Hodel and Jevin West. 2023.
\newblock \href {https://arxiv.org/abs/2308.16118} {Response to {``Emergent analogical reasoning in large language models''}}.
\newblock \emph{arXiv preprint arXiv:2308.16118}.
\newblock Response to Webb et al. (2023), Nature Human Behaviour.

\bibitem[{Hofstadter and Mitchell(1994)}]{hofstadtermitchell1994}
D.~R. Hofstadter and M.~Mitchell. 1994.
\newblock The copycat project: A model of mental fluidity and analogy-making.
\newblock In \emph{Advances in Connectionist and Neural Computation Theory}, volume~2, page 31–112. Ablex, Norwood, NJ.

\bibitem[{Hofstadter(1984)}]{hofstadter1984}
Douglas~R Hofstadter. 1984.
\newblock \href {https://dspace.mit.edu/handle/1721.1/5648} {The {Copycat} project: An experiment in nondeterminism and creative analogies}.
\newblock Technical report, Massachusetts Institute of Technology.

\bibitem[{Holyoak(2012)}]{holyoak2012analogy}
Keith~J Holyoak. 2012.
\newblock Analogy and relational reasoning.
\newblock In K.~J. Holyoak and R.~G. Morrison, editors, \emph{The Oxford handbook of thinking and reasoning}, pages 234--259. Oxford University Press.

\bibitem[{Holyoak et~al.(1984)Holyoak, Junn, and Billman}]{holyoak1984development}
Keith~J Holyoak, Ellen~N Junn, and Dorrit~O Billman. 1984.
\newblock \href {https://doi.org/10.2307/1129778} {Development of analogical problem-solving skill}.
\newblock \emph{Child development}, pages 2042--2055.

\bibitem[{Huang and Chang(2022)}]{huang2022reasoningllmssurvey}
Jie Huang and Kevin Chen-Chuan Chang. 2022.
\newblock \href {https://doi.org/10.48550/arXiv.2212.10403} {Towards reasoning in large language models: A survey}.
\newblock \emph{CoRR}, abs/2212.10403.

\bibitem[{Ichien et~al.(2020)Ichien, Lu, and Holyoak}]{ichien2020}
Nicholas Ichien, Hongjing Lu, and Keith~J Holyoak. 2020.
\newblock \href {https://doi.org/10.3758/s13428-019-01312-3} {Verbal analogy problem sets: An inventory of testing materials}.
\newblock \emph{Behavior Research Methods}, 52(5):1803--1816.

\bibitem[{Johnson et~al.(2025)Johnson, ter Veen, Choenni, van~der Maas, Shutova, and Stevenson}]{johnson2025verbal}
Tamar Johnson, Mathilde ter Veen, Rochelle Choenni, Han van~der Maas, Ekaterina Shutova, and Claire~E Stevenson. 2025.
\newblock Do large language models solve verbal analogies like children do?
\newblock In \emph{Proceedings of the 29th Conference on Computational Natural Language Learning}, pages 627--639.

\bibitem[{Jones et~al.(2022)Jones, Kmiecik, Irwin, and Morrison}]{jones2022}
Laura~L Jones, Matt~J Kmiecik, John~L Irwin, and Robert~G Morrison. 2022.
\newblock \href {https://doi.org/10.3758/s13423-022-02062-8} {Differential effects of semantic distance, distractor salience, and relations in verbal analogy}.
\newblock \emph{Psychonomic Bulletin \& Review}.

\bibitem[{Lewis and Mitchell(2025)}]{lewis2025evaluating}
Martha Lewis and Melanie Mitchell. 2025.
\newblock Evaluating the robustness of analogical reasoning in {GPT} models.
\newblock \emph{Transactions on Machine Learning Research}.

\bibitem[{McCoy et~al.(2024)McCoy, Yao, Friedman, Hardy, and Griffiths}]{mccoy2024embers}
R.~Thomas McCoy, Shunyu Yao, Dan Friedman, Matthew Hardy, and Thomas~L. Griffiths. 2024.
\newblock \href {https://doi.org/10.1073/pnas.2322420121} {Embers of autoregression show how large language models are shaped by the problem they are trained to solve}.
\newblock \emph{Proceedings of the National Academy of Sciences}, 121(41):e2322420121.
\newblock Edited by Richard Shiffrin.

\bibitem[{Mitchell(2021)}]{mitchell2021abstraction}
Melanie Mitchell. 2021.
\newblock Abstraction and analogy-making in artificial intelligence.
\newblock \emph{Annals of the New York Academy of Sciences}, 1505(1):79--101.

\bibitem[{Moskvichev et~al.(2023)Moskvichev, Odouard, and Mitchell}]{moskvichev23}
Arseny Moskvichev, Victor~V. Odouard, and Melanie Mitchell. 2023.
\newblock \href {https://doi.org/10.48550/arXiv.2305.07141} {The conceptarc benchmark: Evaluating understanding and generalization in the arc domain}.
\newblock \emph{arXiv:2305.07141}.

\bibitem[{Mulholland et~al.(1980)Mulholland, Pellegrino, and Glaser}]{mulholland1980components}
Timothy~M Mulholland, James~W Pellegrino, and Robert Glaser. 1980.
\newblock Components of geometric analogy solution.
\newblock \emph{Cognitive psychology}, 12(2):252--284.

\bibitem[{OpenAI(2023)}]{llm_gpt4}
OpenAI. 2023.
\newblock \href {https://cdn.openai.com/papers/gpt-4.pdf} {Gpt-4 technical report.}
\newblock \emph{arXiv preprint arXiv:2303.08774}.

\bibitem[{Opie{\l}ka et~al.(2025)Opie{\l}ka, Rosenbusch, and Stevenson}]{opielka2025analogical}
Gustaw Opie{\l}ka, Hannes Rosenbusch, and Claire~E Stevenson. 2025.
\newblock \href {https://arxiv.org/pdf/2503.03666} {Analogical reasoning inside large language models: Concept vectors and the limits of abstraction}.
\newblock \emph{arXiv preprint arXiv:2503.03666}.

\bibitem[{Opie{\l}ka et~al.(2024)Opie{\l}ka, Rosenbusch, Vijverberg, and Stevenson}]{opielka2024large}
Gustaw Opie{\l}ka, Hannes Rosenbusch, Veerle Vijverberg, and Claire~E Stevenson. 2024.
\newblock Do large language models solve {ARC} visual analogies like people do?
\newblock In \emph{Proceedings of the 46th Annual Conference of the Cognitive Science Society}.

\bibitem[{Razeghi et~al.(2022)Razeghi, Logan~IV, Gardner, and Singh}]{razeghi2022impact}
Yasaman Razeghi, Robert~L Logan~IV, Matt Gardner, and Sameer Singh. 2022.
\newblock \href {https://aclanthology.org/2022.findings-emnlp.59.pdf} {Impact of pretraining term frequencies on few-shot reasoning}.
\newblock In \emph{Findings of the Association for Computational Linguistics: EMNLP 2022}, pages 840--854.

\bibitem[{Richland et~al.(2006)Richland, Morrison, and Holyoak}]{richland2006children}
Lindsey~E Richland, Robert~G Morrison, and Keith~J Holyoak. 2006.
\newblock Children’s development of analogical reasoning: Insights from scene analogy problems.
\newblock \emph{Journal of experimental child psychology}, 94(3):249--273.

\bibitem[{Schank(1980)}]{schank1980aiintelligence}
Roger~C Schank. 1980.
\newblock \href {https://www.sciencedirect.com/science/article/abs/pii/0160289680900021} {How much intelligence is there in artificial intelligence?}
\newblock \emph{Intelligence}, 4:1--14.

\bibitem[{Shiffrin and Mitchell(2023)}]{shiffrin2023probing}
Richard Shiffrin and Melanie Mitchell. 2023.
\newblock Probing the psychology of ai models.
\newblock \emph{Proceedings of the National Academy of Sciences}, 120(10):e2300963120.

\bibitem[{Stevenson and Hickendorff(2018)}]{stevenson2018}
Claire~E Stevenson and Marian Hickendorff. 2018.
\newblock \href {https://doi.org/10.1016/j.lindif.2018.04.010} {Learning to solve figural matrix analogies: The paths children take}.
\newblock \emph{Learning and Individual Differences}, 66:16--28.

\bibitem[{Thibaut and French(2016)}]{thibaut2016analogical}
Jean-Pierre Thibaut and Robert~M French. 2016.
\newblock \href {https://www.sciencedirect.com/science/article/abs/pii/S0885201415300186} {Analogical reasoning, control and executive functions: a developmental investigation with eye-tracking}.
\newblock \emph{Cognitive Development}, 38:10--26.

\bibitem[{Thibaut et~al.(2022)Thibaut, Glady, and French}]{thibaut2022acrossformats}
Jean-Pierre Thibaut, Yannick Glady, and Robert~M French. 2022.
\newblock Understanding the what and when of analogical reasoning across analogy formats: An eye-tracking and machine learning approach.
\newblock \emph{Cognitive Science}, 46(11):e13208.

\bibitem[{Webb et~al.(2023)Webb, Holyoak, and Lu}]{webb2023emergent}
Taylor Webb, Keith~J Holyoak, and Hongjing Lu. 2023.
\newblock \href {https://www.nature.com/articles/s41562-023-01659-w} {Emergent analogical reasoning in large language models}.
\newblock \emph{Nature Human Behaviour}, 7(9):1526--1541.

\bibitem[{Webb et~al.(2024)Webb, Holyoak, and Lu}]{webb2024evidence}
Taylor Webb, Keith~J Holyoak, and Hongjing Lu. 2024.
\newblock \href {https://arxiv.org/abs/2404.13070} {Evidence from counterfactual tasks supports emergent analogical reasoning in large language models}.
\newblock \emph{arXiv preprint arXiv:2404.13070}.

\bibitem[{Wei et~al.(2022)Wei, Tay, Bommasani, Raffel, Zoph, Borgeaud, Yogatama, Bosma, Zhou, Metzler, Chi, Hashimoto, Vinyals, Liang, Dean, and Fedus}]{wei2022emergent}
Jason Wei, Yi~Tay, Rishi Bommasani, Colin Raffel, Barret Zoph, Sebastian Borgeaud, Dani Yogatama, Maarten Bosma, Denny Zhou, Donald Metzler, Ed~H. Chi, Tatsunori Hashimoto, Oriol Vinyals, Percy Liang, Jeff Dean, and William Fedus. 2022.
\newblock \href {https://openreview.net/forum?id=yzkSU5zdwD} {Emergent abilities of large language models}.
\newblock \emph{Transactions on Machine Learning Research}.
\newblock Survey Certification.

\bibitem[{Wu et~al.(2024)Wu, Qiu, Ross, Aky{\"u}rek, Chen, Wang, Kim, Andreas, and Kim}]{wu2024reasoning}
Zhaofeng Wu, Linlu Qiu, Alexis Ross, Ekin Aky{\"u}rek, Boyuan Chen, Bailin Wang, Najoung Kim, Jacob Andreas, and Yoon Kim. 2024.
\newblock \href {https://doi.org/10.18653/v1/2024.naacl-long.102} {Reasoning or reciting? exploring the capabilities and limitations of language models through counterfactual tasks}.
\newblock In \emph{Proceedings of the 2024 Conference of the North American Chapter of the Association for Computational Linguistics}, volume~1 of \emph{NAACL-HLT '24}, pages 1819--1862, Mexico City, Mexico. Association for Computational Linguistics.

\bibitem[{Yiu et~al.(2024)Yiu, Qraitem, Wong, Majhi, Bai, Ginosar, Gopnik, and Saenko}]{yiu2024kiva}
Eunice Yiu, Maan Qraitem, Charlie Wong, Anisa~Noor Majhi, Yutong Bai, Shiry Ginosar, Alison Gopnik, and Kate Saenko. 2024.
\newblock \href {https://arxiv.org/abs/2407.17773} {Kiva: Kid-inspired visual analogies for testing large multimodal models}.
\newblock \emph{arXiv preprint arXiv:2407.17773}.

\end{thebibliography}

\newpage
\onecolumn 
\appendix
\section{Open Science Practices}
\paragraph{Ethics} This study was approved by the ethics board at the University of Amsterdam, Social and Behavioural Sciences on May 24, 2023 (ID: FMG-2495).
\paragraph{Preregistration} The hypotheses, methods and analyses were preregistered on the Open Science Framework (OSF) prior to ethical approval. These were updated on July 26, 2023, to accommodate new methods for LLM data collection. The main deviation from preregistration was using different LLMs than previously specified. 
\paragraph{Data Availability} Preregistration, materials, data, and code are publicly available on the project’s OSF repository: \url{https://osf.io/jdty3/}. A direct link to all data and code can be found here: \url{https://github.com/cstevenson-uva/llm_letterstring_generalization/}.

\section{LLM prompt engineering results}
\label{app:templates}
We administered each letter-string analogy item to LLMs using 5 different prompt templates, as prompt engineering can change the LLMs' performance on the task. The templates were as follows.
\begin{enumerate}
    \itemsep 0em 
    \item If a b c changes to a b d, what does i j k change to?
    \item a b c is to a b d, as i j k is to ?
    \item{a b c → a b d \textbackslash n e f g → }
    \item{Let's try to complete the pattern:\textbackslash n [a b c] [a b d] \textbackslash n [i j k] [}
    \item{[a b c] [a b d] \textbackslash n [i j k] [}
\end{enumerate}

As can be seen in Figure or Table \ref{tbl:templates}, template 1, derived from \cite{mitchell2021abstraction} worked best overall. Template 4, the best template found by \cite{webb2023emergent} worked well in Latin and Greek alphabets, but not as well for the Symbol list, which makes sense because $[$ and $]$ are symbols themselves. Our results are based on template 1.

\begin{table*}[h]
\centering
\caption{\fontsize{10}{12}\selectfont Prompt Template Performance Mean Correct (SE) for Selected Models}
\label{tbl:templates}
{\fontsize{10}{12}\selectfont
\begin{tabular}{lccccc}
\toprule
\textbf{Model} & \textbf{Template 1} & \textbf{Template 2} & \textbf{Template 3} & \textbf{Template 4} & \textbf{Template 5} \\
\midrule
Claude-3.5     & 0.82 (0.10) & \textbf{0.88 (0.08)} & 0.71 (0.11) & 0.53 (0.13) & 0.71 (0.11) \\
Gemma-2 27B    & \textbf{0.59 (0.12)} & \textbf{0.59 (0.12)} & 0.41 (0.12) & 0.41 (0.12) & 0.29 (0.11) \\
GPT-4o         & \textbf{0.82 (0.10)} & 0.71 (0.11) & 0.71 (0.11) & 0.71 (0.11) & 0.71 (0.11) \\
Llama-3.1 405B & \textbf{0.71 (0.11)} & 0.59 (0.12) & 0.59 (0.12) & 0.59 (0.12) & 0.35 (0.12) \\
\midrule
Total          & \textbf{0.74 (0.05)} & 0.69 (0.06) & 0.60 (0.06) & 0.56 (0.06) & 0.52 (0.06) \\
\bottomrule
\end{tabular}}
\end{table*}

\section{LLM results without previous messages}
\label{app:noprevmsg}
We readministered the items from the template comparison (see Appendix \ref{app:templates}) to examine whether it was better to administer the items one-by-one or to include all previous message history, i.e. the previous items and their responses. 

As can be seen in Table \ref{tbl:prevmsg}, it was generally advantageous to include previous message history versus not. Of the LLMs we tested, there may be two possible exceptions to look out for in future work. Both Gemma-2 27B and Llama-3.1 405B had significantly higher accuracy (i.e., no overlapping SE margins) without message history on the Symbol alphabet. In both cases, the main result of lower performance on Greek and Symbol alphabets versus Latin alphabet still holds.

\begin{table*}[h]
\centering
\caption{{\fontsize{10}{12}\selectfont Mean Correct (SE) for LLMs with versus without Message History.}}
\label{tbl:prevmsg}
{\fontsize{10}{12}\selectfont
\begin{tabular}{lcc|cc|cc}
\toprule
\textbf{Model} & \multicolumn{2}{c}{\textbf{Latin}} & \multicolumn{2}{c}{\textbf{Greek}} & \multicolumn{2}{c}{\textbf{Symbol}} \\
& \textbf{Incl. History} & \textbf{No History} & \textbf{Incl. History} & \textbf{No History} & \textbf{Incl. History} & \textbf{No History} \\
\midrule
Claude-3.5     & \textbf{0.88 (0.07)} & 0.65 (0.10) 
               & \textbf{0.80 (0.08)} & 0.75 (0.10) 
               & \textbf{0.40 (0.10)} & 0.20 (0.09) \\
Gemma-2 27B    & 0.64 (0.10) & \textbf{0.73 (0.08)} 
               & \textbf{0.48 (0.10)} & 0.43 (0.09) 
               & 0.04 (0.04) & \textbf{0.20 (0.07)} \\
GPT-4o         & \textbf{0.84 (0.07)} & 0.73 (0.08) 
               & \textbf{0.64 (0.10)} & 0.43 (0.09) 
               & \textbf{0.60 (0.10)} & 0.57 (0.09) \\
Llama-3.1 405B & \textbf{0.76 (0.09)} & 0.67 (0.09) 
               & 0.56 (0.10) & \textbf{0.67 (0.09)} 
               & 0.20 (0.08) & \textbf{0.43 (0.09)} \\
\bottomrule
\end{tabular}}
\end{table*}

\section{Ordered Symbol Task}
\label{app:orderedsymboltask}
We re-administered the items from the template comparison (see Appendix \ref{app:templates}) to examine whether ordering the symbols by unicode value would improve the models' performance on the Symbol alphabet. This did result in improved performance in Claude 3.5 and Gemma 2, where Claude 3.5 improved to the same performance level of the Greek alphabet. For GPT-4o and Llama 3.1 there were no significant improvements from the reordering. In all cases our main finding \textemdash that performance degraded from Latin to the Greek and Symbol alphabets \textemdash still held. However, in future experiments using a Symbol domain, it is important to realize that LLMs generally benefit by ordering symbols by unicode value.

\begin{table*}[h]
\caption{{\fontsize{10}{12}\selectfont Mean (SE) correct for LLMs with versus without Symbols were ordered by unicode value.}}
\centering
{\fontsize{10}{12}\selectfont
\begin{tabular}{lcc|cc}
\toprule
\textbf{Model} & \textbf{Latin} & \textbf{Greek} & \textbf{Symbol (unordered)} & \textbf{Symbol (ordered)} \\
\midrule
Claude-3.5     & 0.84 (0.07) & 0.72 (0.09) & 0.40 (0.10) & \textbf{0.72 (0.09)} \\
Gemma-2 27B    & 0.64 (0.10) & 0.48 (0.10) & 0.04 (0.04) & \textbf{0.36 (0.10)} \\
GPT-4o         & 0.76 (0.09) & 0.60 (0.10) & \textbf{0.60 (0.10)} & \textbf{0.60 (0.10)} \\
Llama-3.1 405B & 0.72 (0.09) & 0.68 (0.10) & \textbf{0.20 (0.08)} & \textbf{0.28 (0.09)} \\
\midrule
Total & 0.74 (0.44) & 0.62 (0.49) & 0.31 (0.47) & \textbf{0.49 (0.50)} \\
\bottomrule
\end{tabular}}
\label{tbl:effectorderedsymbols}
\end{table*}

\end{document}